\documentclass[conference]{IEEEtran}
\usepackage[utf8]{inputenc}
\usepackage{amsmath}
\usepackage{graphicx}
\usepackage{cite}
\usepackage{url}
\usepackage{hyperref}
\usepackage{booktabs}
\usepackage{array}
\usepackage{multirow}
\usepackage{makecell}
\usepackage{threeparttable}
\usepackage{tabularx}

\usepackage{amsmath,amsfonts,amssymb}

\usepackage{listings}
\makeatletter
\lst@AddToHook{OnEmptyLine}{\vspace{-4pt}}
\makeatother

\makeatletter
\newcommand{\newlineauthors}{%
  \end{@IEEEauthorhalign}\hfill\mbox{}\par
  \mbox{}\hfill\begin{@IEEEauthorhalign}
}
\makeatother

\title{Auto-ML Graph Neural Network Hypermodels for Outcome Prediction in Event-Sequence Data}

\author{
\IEEEauthorblockN{Fang Wang}
\IEEEauthorblockA{Khalifa University\\Computer Science Department\\Abu Dhabi, United Arab Emirates\\florence.wong@ku.ac.ae}
\and
\IEEEauthorblockN{Lance Kosca}
\IEEEauthorblockA{Khalifa University\\ Mechanical and Nuclear Engineering Department\\Abu Dhabi, United Arab Emirates\\lance.kosca@ku.ac.ae}
\and
\IEEEauthorblockN{Adrienne Kosca}
\IEEEauthorblockA{De La Salle University\\Engineer Department\\Manila, Philippines\\adriennekosca@gmail.com}
\newlineauthors
\IEEEauthorblockN{Marko Gacesa}  
\IEEEauthorblockA{Khalifa University\\Physics Department\\Abu Dhabi, United Arab Emirates\\marko.gacesa@ku.ac.ae}
\and
\IEEEauthorblockN{Ernesto Damiani}
\IEEEauthorblockA{University of Milan\\Computer Science Department\\Milan, Italy\\ernesto.damiani@unimi.it}
}

\begin{document}

\maketitle

\begin{abstract}
This paper introduces HGNN(O), an AutoML GNN hypermodel framework for outcome prediction on event-sequence data. Building on our earlier work on graph convolutional network hypermodels, HGNN(O) extends four architectures—One Level, Two Level, Two Level Pseudo Embedding, and Two Level Embedding—across six canonical GNN operators. A self-tuning mechanism based on Bayesian optimization with pruning and early stopping enables efficient adaptation over architectures and hyperparameters without manual configuration. Empirical evaluation on both balanced and imbalanced event logs shows that HGNN(O) achieves accuracy exceeding 0.98 on the \textit{Traffic Fines} dataset and weighted F1 scores up to 0.86 on the \textit{Patients} dataset without explicit imbalance handling. These results demonstrate that the proposed AutoML–GNN approach provides a robust and generalizable benchmark for outcome prediction in complex event-sequence data.

\end{abstract}


\section{Introduction}
Event-sequence data, such as business process logs, record the temporal evolution of activities and their attributes. These data are increasingly exploited for predictive analytics tasks. Among them, outcome prediction \cite{teinemaa2019outcome} is a central objective: forecasting the final state of an ongoing case based on its observed prefix. Accurate outcome prediction enables proactive interventions, supports compliance, and improves operational efficiency, making it a core problem in predictive business process monitoring (PBPM). First, outcome prediction must generalize across diverse datasets, which motivates hypermodel frameworks that can adapt architectures and parameters automatically. Second, event-sequence data exhibit complex temporal and relational dependencies that are difficult to capture with purely sequential models, highlighting the need for graph-based representations and GNNs \cite{wang2025time}.

To address these challenges, we introduce HGNN(O), an AutoML GNN hypermodel framework for outcome prediction on event sequences. The current model builds on our earlier work on hypermodel graph convolutional networks\cite{wang2025hgcn}, which introduced four architectures: One Level, Two Level, Two Level Pseudo Embedding, and Two Level Embedding. In this paper, these architectures are generalized to a broader AutoML GNN framework.

The contributions of this work are threefold. First, we present a unified framework that integrates these four architectures with six canonical GNN operators, namely GCNConv\cite{kipf2016semi}, GraphConv\cite{morris2019weisfeiler}, SAGEConv\cite{hamilton2017inductive}, TAGConv\cite{du2017topology}, ChebConv\cite{defferrard2016convolutional}, and GINConv\cite{xu2018powerful}, enabling systematic exploration of combinations between operator and architecture. This framework also serves as a benchmark for evaluating the relative effectiveness of different GNN designs in predictive business process monitoring. Second,we introduce a self-tuning mechanism based on Bayesian hyperparameter optimization with pruning and early stopping, which ensures adaptability across diverse event-sequence datasets by searching effectively across architectures and hyperparameters. Third, we conduct an empirical evaluation on both balanced and imbalanced event logs, demonstrating the robustness and generalizability of the proposed approach.

The remainder of this paper is organized as follows. Section 2 reviews related work, and Section 3 introduces the graph representation and GNN hypermodels. Section 4 describes the datasets and experimental setup, Section 5 reports the results, and Section 6 concludes the paper.

\section{Related Work}
Early approaches to predictive business process monitoring (PBPM) relied on symbolic or feature-based sequence classification. In these methods, machine-learning models were trained on hand-crafted attributes derived from event logs \cite{leontjeva2015complex}. Commonly used algorithms included Decision Trees (DT) \cite{de2016general}, Random Forests (RF) \cite{leontjeva2015complex}, XGBoost \cite{teinemaa2019outcome}, and Support Vector Machines (SVMs) \cite{kang2012periodic}. Although ensemble and boosting techniques often performed well on structured tabular data, they struggled to model the temporal order and contextual dependencies between events, which limited their ability to generalize across real-world process traces \cite{teinemaa2019outcome}. For instance, the SVM-based method in \cite{kang2012periodic} achieved an accuracy of 82\% in controlled experiments but declined to 63\% when evaluated on real hospital data \cite{rama2021deep}.

To overcome the rigidity of manual feature construction, sequence modeling techniques based on Recurrent Neural Networks (RNNs), particularly Long Short-Term Memory (LSTM) networks, were adopted for PBPM \cite{rama2021deep}. Early works mainly addressed next-event prediction and remaining time estimation tasks \cite{tax2017predictive}, while later studies extended these models to handle richer, multi-attribute event encodings \cite{lin2019mm}. Nevertheless, LSTM architectures continue to face structural limitations in modeling complex temporal dependencies and maintaining stable performance across heterogeneous process types. Convolutional Neural Networks (CNNs) \cite{pasquadibisceglie2020orange} have also been explored to extract local sequential features more effectively, but their fixed receptive fields restrict their ability to capture long-range interactions and higher-order event dependencies present in process sequences.

Recent research has therefore shifted toward graph-based approaches, where event logs are represented as graphs to better capture relational and temporal structures. Graph Neural Networks (GNNs) \cite{wang2025time} offer a flexible framework to model these interactions, enabling richer feature propagation and more interpretable representations. However, most GNN-based methods rely on manually defined architectures and hyperparameters, reducing their generalizability across heterogeneous datasets.

\section{Methodology}
\subsection{Graph Representation of Event-Sequence Data}
We formulate predictive business process monitoring as a classification task, where the goal is to learn a mapping function from a sequence of events (trace) to a categorical outcome label. Each sequence is modeled as a directed, weighted graph $\mathcal{G}_{G_j}$, where nodes correspond to events and edges represent their temporal ordering. For every node $X_i$, we construct a feature vector $\mathbf{v}_{N_i} = [A_i, U_i, B_i]$ that combines the activity label $A_i$, universal attributes $U_i$, and event-specific attributes $B_i$. Categorical features are represented by one-hot encoding, while numerical features are normalized using min–max scaling. Missing values are imputed by the median for numerical attributes and by a padding token ($-1$) for categorical attributes.
At the sequence level, each graph $G_j$ is associated with a global feature vector $\mathbf{v}_{G_j}$. The full graph is defined by a node matrix $\mathbb{V}_{N(G_j)} \in \mathbb{R}^{n \times d_N}$, an edge index tensor $\mathbb{E}_{G_j} \in \mathbb{R}^{2 \times (n-1)}$ linking consecutive nodes, and an edge weight vector $\mathbb{W}_{G_j} \in \mathbb{R}^{n-1}$. Edge weights are normalized time gaps between event start times, $w_{(i \rightarrow i+1)} = T^s_{i+1} - T^s_i$, which preserve causality ($w=0$ for simultaneous events). 

\subsection{GNN Hypermodels}
HGNN(O) constructs four hypermodel architectures using six foundational GNN layers: GCNConv\cite{kipf2016semi}, GraphConv\cite{morris2019weisfeiler}, SAGEConv\cite{hamilton2017inductive}, TAGConv\cite{du2017topology}, ChebConv\cite{defferrard2016convolutional}, and GINConv\cite{xu2018powerful}. These operators were selected because they represent widely used GNN families, covering spectral and spatial paradigms as well as key design variations: localized spectral filters (ChebConv, GCNConv), neighborhood aggregation (GraphConv, SAGEConv), temporal-aware diffusion (TAGConv), and expressive Weisfeiler–Lehman mappings (GINConv). Together, they provide a diverse and representative basis for benchmarking. 

\subsubsection{Foundational GNN Operators}
To formalize their role within our hypermodels, we summarize the canonical update rules of these operators. Each instantiates the message-passing paradigm with distinct aggregation or filtering strategies, capturing complementary design choices across spectral and spatial GNNs. We denote node feature matrices by $\mathbb{V}^{(l)}$ and per-node embeddings by $\mathbf{h}_v^{(l)}$.

We employ a weighted extension of the GCNconv layer, modifying the original formulation\cite{kipf2016semi} in to account for temporal edge information. At propagation step $l$, node representations are updated by
\begin{equation}
\mathbb{V}^{(l+1)} = \sigma \!\left( \hat{D}^{-\tfrac{1}{2}} \hat{A}_w \hat{D}^{-\tfrac{1}{2}} \mathbb{V}^{(l)} W^{(l)} \right),
\end{equation}
where $\hat{A}_w = A_w + I$ augments the weighted adjacency with self-connections, and $\hat{D}$ is its degree matrix. $\sigma(\cdot)$ denotes a nonlinear activation and $W^{(l)}$ is the learnable transformation. The weights $A_w$ correspond to normalized inter-event time gaps, so the convolution integrates both structural and temporal dependencies. 

Following\cite{morris2019weisfeiler}, the weighted GraphConv layer updates node features as
\begin{equation}
\mathbb{V}^{(l+1)} = \sigma \!\left(D_w^{-1} A_w \mathbb{V}^{(l)} W^{(l)} \right),
\end{equation}
where $A_w$ is the weighted adjacency and $D_w$ its degree matrix. Unlike GCNConv, it applies left normalization only, so neighbor messages are scaled directly by inverse degree. 

We implement the mean-aggregator variant of GraphSAGE, where node features are updated as
\begin{equation}
\mathbf{h}_v^{(l+1)} = \sigma \!\left( W^{(l)} \cdot 
\left[ \mathbf{h}_v^{(l)} \,\|\, \frac{1}{|\mathcal{N}(v)|}\!\!\sum_{u \in \mathcal{N}(v)} \mathbf{h}_u^{(l)} \right] \right),
\end{equation}
with $\mathcal{N}(v)$ the neighbors of $v$, $\|\,$ denoting concatenation, $W^{(l)}$ the learnable weights, and $\sigma$ a nonlinearity. 

Temporal Graph Convolution (TAGConv) applies a truncated polynomial filter over powers of the adjacency:
\begin{equation}
\mathbb{V}^{(l+1)} = \sigma \!\left( \sum_{k=0}^{K} (A_w^k \mathbb{V}^{(l)} \Theta_k) \right),
\end{equation}
where $K$ is the filter size and $\{\Theta_k\}$ are learnable weights. This captures multi-hop temporal diffusion through successive adjacency powers. 

Chebyshev convolution (ChebConv) uses polynomial filters of the normalized Laplacian:
\begin{equation}
\mathbb{V}^{(l+1)} = \sum_{k=0}^{K} T_k(\tilde{L}) \,\mathbb{V}^{(l)} \Theta_k,
\end{equation}
where $\tilde{L}$ is the rescaled Laplacian, $T_k$ are Chebyshev polynomials, and $\{\Theta_k\}$ are learnable coefficients. This provides localized spectral filtering up to $K$ hops.

The Graph Isomorphism Network (GIN) updates node features via
\begin{equation}
\mathbf{h}_v^{(l+1)} = \mathrm{MLP}^{(l)} \!\left( (1+\epsilon)\mathbf{h}_v^{(l)} + \sum_{u \in \mathcal{N}(v)} \mathbf{h}_u^{(l)} \right),
\end{equation}
where $\epsilon$ is a learnable (or fixed) scalar. This design maximizes discriminative power, matching the Weisfeiler–Lehman test.

\subsubsection{GNN Architectures}
Each hypermodel integrates one of the six GNN operators as its core layer. The surrounding architecture varies to capture structural and temporal dependencies, particularly by fusing event-level and sequence-level features with node and edge attributes.

\textbf{O-GNN (One-Level)} augments each node with a broadcasted graph-level vector, enabling joint modeling of local and global features. The enriched graph is processed by GNN layers, pooled into a graph embedding, and classified through fully connected layers. 

\textbf{T-GNN (Two-Level)} processes node-level features through GNN layers and pools them into a graph embedding, while sequence-level attributes are independently transformed by dense layers. The resulting embeddings are concatenated before the final classification. 

\textbf{TP-GNN (Two-Level Pseudo-Embedding)} extends T-GNN by incorporating a pseudo-embedding matrix, processed through its own GNN stack and fused with node-based representations. Here, we use a duration-bin pseudo-embedding matrix\cite{wang2025hgcn,wang2025comprehensive} to provide an alternative feature space that captures temporal patterns not directly observable from raw attributes. 

\textbf{TE-GNN (Two-Level Embedding)} emphasizes the role of a decisive attribute (e.g., activity label $A$) by embedding it separately into a dense vector space, forming an auxiliary graph object processed in parallel with the main node features. This ensures its semantic contribution is explicitly preserved.

\subsection{Hypermodel Configurations}
Hyperparameters were tuned using Bayesian optimization (Optuna), with ranges listed in Table~\ref{tab:hyperparameters}. Common hyperparameters across all models include the number of GNN and fully connected layers, hidden units, activation functions, pooling strategies, and optional mechanisms such as batch normalization, dropout, and L1 regularization. Additional parameters include learning rate schedules, optimizer choice with associated settings, batch size, and loss functions. Model-specific search dimensions were also considered: for GraphConv, we optimize neighborhood aggregation due to the absence of symmetric normalization, while in TE-GNN variants, the embedding dimension of the key attribute is tuned. Across all models, masked values (encoded as $-1$) are explicitly handled during training to mitigate biases introduced by irrelevant node attributes.
 
\begin{table}[htb!]
\begin{threeparttable}
\centering
\caption{Hyperparameters and Their Tuning Ranges/Types}
\label{tab:hyperparameters}
\begin{tabularx}{\linewidth}{l@{\hspace{1pt}}X}
\hline
\small
\textbf{Hyperparameter}                          & \textbf{Range}                            \\\toprule
\multicolumn{2}{p{\hsize}}{\textit{\textbf{GNN Layers}}}
\\
Number of layers     & 1-5   \\                                
Hidden Units                    & 16-512 \\
Skip Connection & Y/N\\
Dropout      & flag: Y/N; rates: 0.2-0.7\\
Batch norm & flag: Y/N; momentum: 0.1-0.999; eps:1e-5-1e-2 \\ 
Activation & ReLU, Leaky\_ReLU, ELU, Tanh, Softplus, GELU\\
GraphConv Aggr & add, mean, max \\\midrule
\textit{\textbf{Pooling Method}} & mean, add, max\\\midrule
\multicolumn{2}{p{\hsize}}{\textit{\textbf{Fully Connected Layers}}}\\
Number of layers     & 1-3   \\                                
Dense Units                    & 16-512 \\
Dropout      & flag: Y/N; rates: 0.2-0.7\\
Batch norm & flag: Y/N; momentum: 0.1-0.999; eps:1e-5-1e-2 \\ 
Activation & ReLU, Leaky\_ReLU, ELU, Tanh, Softplus, GELU\\\midrule
\multicolumn{2}{p{\hsize}} {\textit{\textbf{Optimizer and Learning Rate Scheduler}}}\\
\textbullet\space   Optimizer                    &  \\ 
Learning Rate & 1e-5-1e-2 (log)\\
Weight Decay & 0-1e-3\\
L1 & 0-1e-3\\
\multicolumn{2}{p{\hsize}}{Type of Optimizers}\\
\textit{Adam}                 & $\beta_1$: 0.85-0.99; $\beta_2$:0.99-0.999 \\ 
\textit{SGD}                 & momentum: 0.0-0.9  \\ 
\textit{RMSprop}& $\alpha$: 0.9-0.999; momentum: 0.0-0.9; eps: 1e-9-1e-7 \\

\multicolumn{2}{p{\hsize}}{\textbullet\space Learning Rate Schedulers}
\\
\textit{Step}       &  step size: 1-50; $\gamma$: 0.1-0.9\\
\textit{Exponential} & $\gamma$: 0.85-0.99\\
\textit{Reduce-on-Plateau} & factor: 0.1-0.9; patience: 1-50; threshold: 1e-4-1e-2; eps: 1e-8-1e-4 \\
\textit{Polynomial} & power: 0.1-2; total\_iters: 2-300\\ 
\textit{Cosine Annealing} & T\_max: 10-100; eta\_min: 1e-6-1e-2\\
\textit{Cyclic} & base: 1e-5-1e-2 (log); max: 1e-3-1e-1 (log); step\_size\_up: 5-200\\
\textit{One Cycle} & max: 1e-3-1e-1; total\_steps: batch\_size*1000; pct\_start: 0.1-0.5 \\ \midrule
\textit{\textbf{Loss Function}} & CrossEntropy, MultiMargin\\\midrule
\textit{\textbf{Batch Size}} & 16, 32, 64, 128, 512 \\\midrule
\textit{\textbf{Embedding Dim}} & 10-50 \\\bottomrule
\end{tabularx}
    \begin{tablenotes}
	\item{\textit{Note: The abbreviations for parameters and hyperparameters in this table are derived from the PyTorch library, such as ``eps'' for epsilon, ensuring clarity and consistency in terminology.}}
    \end{tablenotes}
\end{threeparttable}
\end{table}

\section{Experiments}
\subsection{Datasets}
The toolkit was evaluated on one balanced and one imbalanced event log dataset. \textbf{Traffic Fines.} This real-life event log \cite{finedataset} contains 9,179 cases from an information system managing road traffic fines. The prediction task involves two outcomes (deviant vs. regular). Sequence-level attributes include two categorical and one numerical feature, while event-level attributes comprise four categorical and two numerical attributes. Cases are balanced across the two outcome classes. \textbf{Patients.} This synthetic healthcare log comprises 2,142 cases, each representing patient interactions within a healthcare system. The dataset is strongly imbalanced across six outcome classes (majority 40.74\%, minority 1.12\%, ~36:1). Graph-level attributes include three numerical and one categorical feature, while node-level attributes consist of one universal categorical, three numerical, and one event-specific categorical feature. All events in both datasets are timestamped with start and completion times. Event durations are computed as the difference between these times, rounded to the nearest minute. In \textit{Patients}, durations shorter than five minutes are placed into unique bins, while longer durations are grouped into 24 quantile-based bins. In \textit{Traffic Fines}, durations of exactly 1,440 minutes (one day) are assigned to a unique bin, while longer durations are grouped into four quantile-based bins.

\subsection{Experimental Setup}
Experiments were conducted on a Dell Latitude 7450 workstation with an Intel® Core™ Ultra 7 165U CPU (12 cores, 14 threads, 2.1 GHz) and 16 GB RAM, running Windows 11 Pro. Each model configuration was optimized over 200 trials, with a maximum of 300 training epochs per trial. An 80/20 train–validation split was used, and early stopping (patience of 30 epochs) together with pruning was applied to reduce overfitting and computational cost. The best hyperparameter set was selected from the top-performing trial, and models were then retrained for 300 full epochs to assess generalization. For balanced datasets, model ranking was based on test accuracy, with ties broken by test loss and variance; for imbalanced datasets, weighted F1 served as the primary criterion, followed by test loss and variance.

\section{Results}
\subsection{Balanced Dataset}
\begin{figure}[htbp]
\centering
\includegraphics[width=0.45\textwidth]{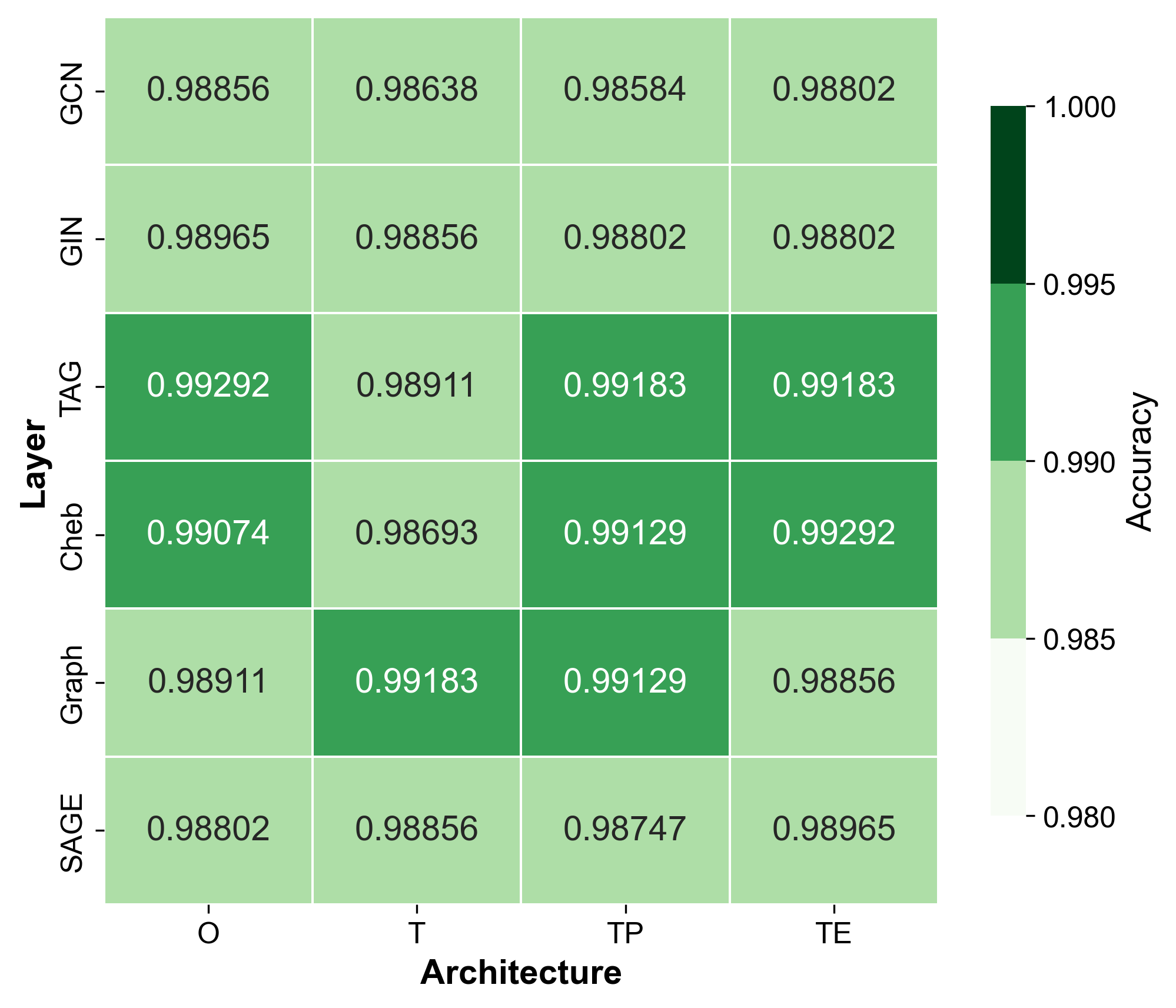}
\caption{Accuracy Heatmap from Training on Traffic Fines Balanced Dataset.}
\label{fig:heatmap_balanced}
\end{figure}

On the balanced \textit{Traffic Fines} log, all hypermodel variants achieved high predictive performance, with test accuracy between 0.986 and 0.993 (see Fig.~\ref{fig:heatmap_balanced}). This confirms the effectiveness of our AutoML–GNN framework: Bayesian optimization with early stopping and pruning consistently identified near-optimal configurations across operators and architectures. Importantly, this success is not due to AutoML alone. The search is effective because the graph representation encodes temporal gaps, event durations, and attribute structures, making the outcome prediction task highly learnable. With this foundation, different GNN operators converge to similarly strong solutions once tuned. Thus, the results highlight the complementary strengths of automated search and graph-based modeling: AutoML ensures efficient exploration of hyperparameters, while the representation provides the expressive basis that makes tuning successful.  

Even the simplest One-Level architecture remained highly competitive across operators, indicating that added complexity from two-level processing or specialized embeddings yields limited benefit in balanced tasks. This has practical importance: in deployment scenarios where efficiency and simplicity matter, lightweight models can achieve accuracy comparable to more elaborate designs. Moreover, the optimization process consistently produced strong configurations across layers and architectures, showing that solutions were not overfitted to dataset idiosyncrasies. In balanced settings, the main advantage of the AutoML–GNN framework lies less in identifying a uniquely superior model than in reliably delivering effective ones without the risk of suboptimal manual choices. This reliability is a key strength for real-world predictive monitoring applications.

For reproducibility, an example of the complete hyperparameter configuration discovered during the search on the balanced \textit{Traffic Fines} dataset is reported in Appendix~\ref{appendix:hyperparameters}.

\subsection{Imbalanced Dataset}
\begin{figure}[htbp]
\centering
\includegraphics[width=0.45\textwidth]{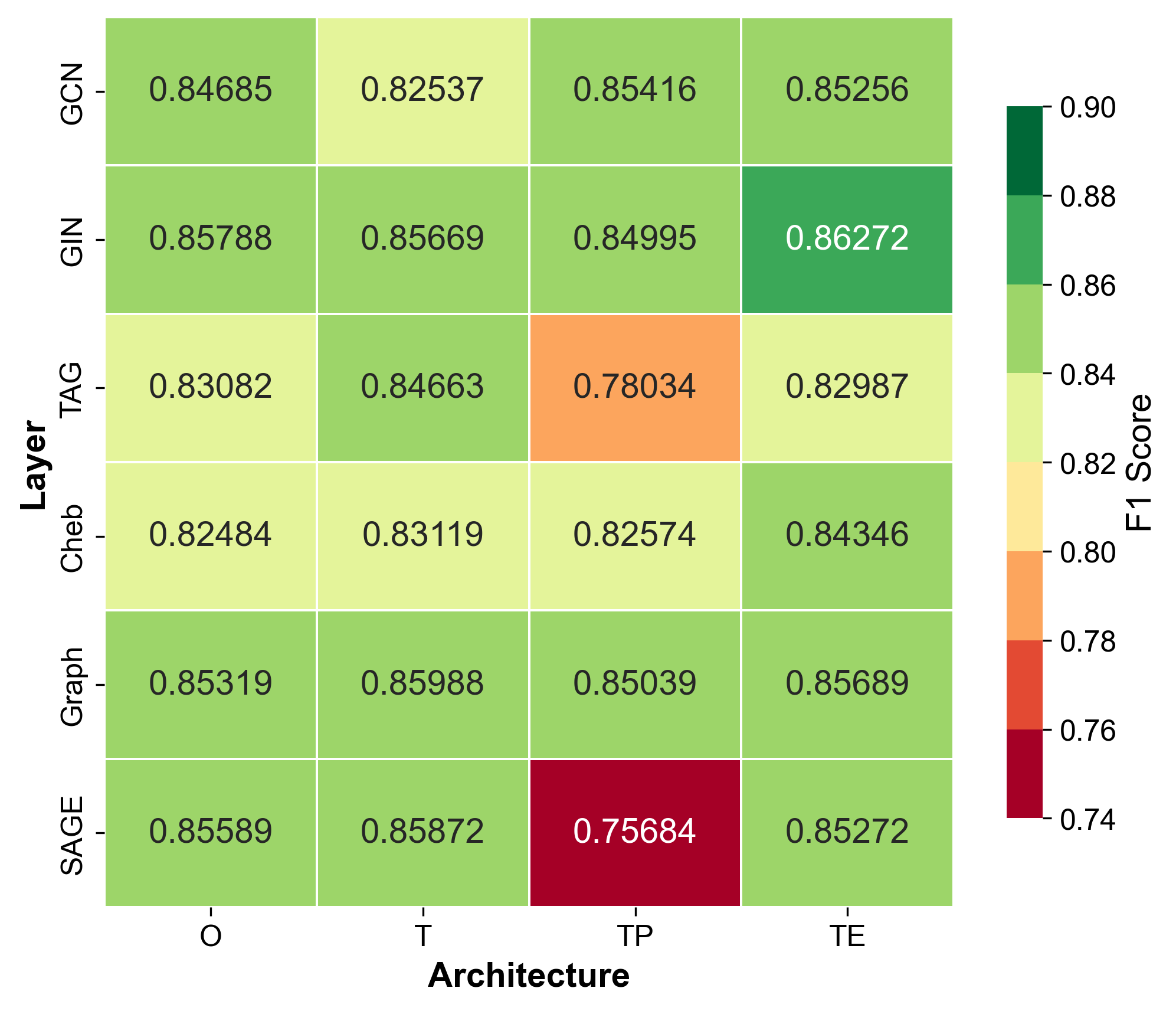}
\caption{F1 Score Heatmap from Training on Patients Imbalanced Dataset.}
\label{fig:heatmap_imbalanced}
\end{figure}

\begin{figure}[htbp]
\centering
\includegraphics[width=0.45\textwidth]{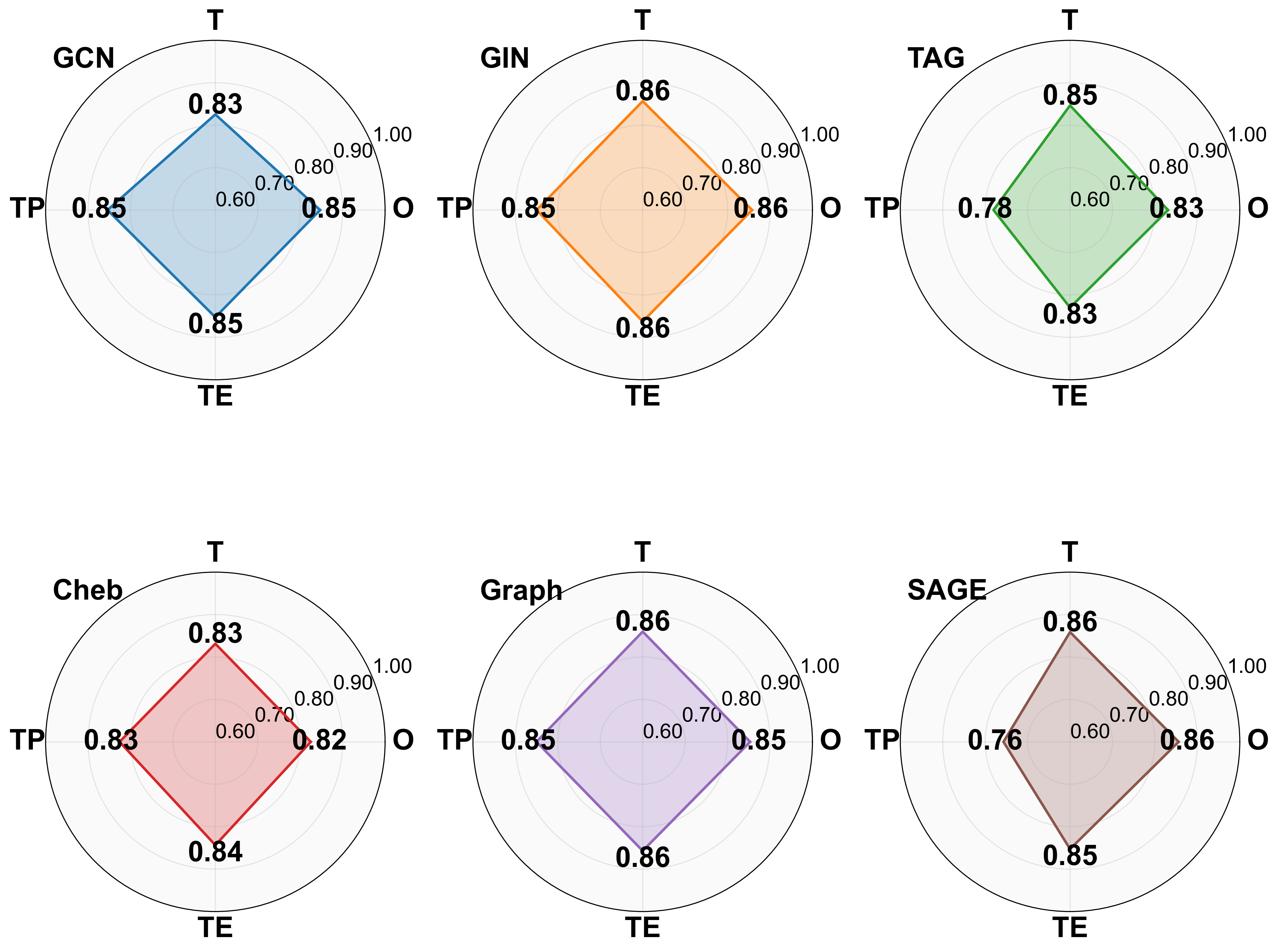}
\caption{F1 Score Radar Charts showing predictive performance of each GNN.}
\label{fig:radar_charts}
\end{figure}

On the imbalanced \textit{Patients} log, models achieved weighted F1 scores between 0.76 and 0.87, as shown in Figs.~\ref{fig:heatmap_imbalanced} and \ref{fig:radar_charts}. This range represents solid predictive performance for a multi-class setting with a ~36:1 imbalance ratio, especially since HGNN(O) does not employ explicit rebalancing strategies. The results therefore highlight the capacity of the AutoML–GNN framework to extract useful patterns from highly skewed data without requiring manual intervention. In the context of healthcare event logs, where rare but critical outcomes are often underrepresented, such resilience is particularly important. It suggests that the framework can deliver clinically meaningful predictions even when faced with the class imbalance that commonly occurs in real-world medical data.

A central finding is the demonstrable robustness of certain GNN layers. GINConv and GraphConv maintained strong performance across nearly all architectural variants, suggesting that they are naturally well suited to capturing the relational dependencies in event log data. In contrast, TAGConv and ChebConv showed more variability, with their effectiveness heavily dependent on surrounding architectural choices. This variability underscores the risk of manual operator selection in imbalanced settings and the importance of automated search in identifying stable configurations. Architectural choices also had a clear effect. Two-Level and Embedding variants generally achieved the strongest results, with the Two-Level model paired with GraphConv producing the best overall performance. This supports the view that explicitly modeling node- and graph-level features separately enhances representational capacity for imbalanced outcome prediction.

In summary, the evaluation on the \textit{Patients} log shows that imbalance exposes clearer differences between operators and architectures than in the balanced setting. While certain layers and designs are more reliable than others, the AutoML–GNN framework consistently identified competitive solutions without requiring handcrafted adjustments. This demonstrates that the proposed approach is not only effective on straightforward balanced tasks but also resilient under the more challenging conditions of skewed medical data, making it a dependable choice for real-world predictive monitoring.

\subsection{Computational Costs}
While HGNN(O) achieves strong predictive performance, this comes with an upfront cost: Bayesian hyperparameter optimization requires training hundreds of candidate configurations. This makes the search process the dominant contributor to runtime. However, the expense is incurred only once, and the outcome is a compact, tuned model that can be deployed with much lower inference costs than would result from manual trial-and-error tuning.  

The efficiency of the final models also varies across operators. For example, GCNConv often reached optimal accuracy with fewer layers and parameters, whereas GraphConv required deeper configurations to achieve similar results. Such variability highlights the value of automated search in balancing predictive power against computational effort.  

Overall, HGNN(O) represents a calculated trade-off: a substantial but automated tuning phase that yields efficient production models, avoiding risks of suboptimal manual choices. Future work will explore multi-fidelity optimization and early termination strategies to further reduce this initial cost.

\section{Conclusions}
This work introduced HGNN(O), an AutoML–GNN framework for outcome prediction of event sequence data that integrates multiple hypermodel architectures and six canonical GNN operators. By combining automated Bayesian optimization with graph representations that capture temporal gaps, event durations, and attribute dependencies, the framework consistently identified high-performing models across both balanced and imbalanced event logs. Experiments showed accuracy exceeding 0.98 on the balanced \textit{Traffic Fines} dataset and strong weighted F1 scores on the imbalanced \textit{Patients} log, despite the absence of explicit rebalancing strategies. These findings demonstrate that the proposed approach delivers both predictive strength and robustness across diverse settings.

Our empirical evaluation supports three main conclusions. First, encoding process traces as graphs with temporally weighted edges and node and graph level attributes provides a powerful representational basis for predictive monitoring. Second, the self-tuning hypermodel design is essential for navigating the architectural and parametric complexity of GNNs and for adapting effectively to both balanced and imbalanced data. Third, among the operators evaluated, GINConv consistently emerged as a stable and reliable choice, particularly when paired with the activity-embedding architecture in the imbalanced setting.

In conclusion, HGNN(O) establishes a strong benchmark for automated, graph-based predictive business process monitoring. Future work will extend the evaluation to larger and more diverse datasets, particularly industrial and IoT logs with extreme imbalance, to further assess generalizability. Reducing search overhead through multi-fidelity or surrogate-assisted optimization, and integrating explicit imbalance-handling strategies such as focal loss or oversampling, represent promising directions for improving efficiency and minority-class performance. Finally, enhancing interpretability through attention based visualization or feature attribution methods will be key for ensuring that the framework remains both effective and practically deployable in real-world settings.

\bibliographystyle{splncs03_unsrt}
\bibliography{ref}

\appendices
\section{Example Hyperparameter Configuration}
The following listing presents the full set of hyperparameters and architectural details of the TAGConv-based TP-GNN model, as identified by HGNN(O) in a representative run on the balanced \textit{Traffic Fines} dataset.
\label{appendix:hyperparameters}
\vspace{0.5em}
\begin{lstlisting}
Output Size: 2
Best batch size: 128
Best epoch: 12
Best accuracy: 0.9858
Best loss: 0.0684
Best loss std: 0.2745

Best hyperparameters found were:
Event Input Size: 24
 Number of GCN layers: 3
   GCN Layer 1:
    Units: 216
    Activation: relu
   GCN Layer 2:
    Units: 155
    Activation: softplus
    Dropout: 0.4965
   GCN Layer 3:
    Units: 232
    Batch Norm Momentum: 0.5215
    Batch Norm Epsilon: 6.3114e-03
    Activation: gelu
    Dropout: 0.2912
 
Duration Embedding Input Size: 10
 Number of Duration Embedding GCN layers: 2
   Duration Embedding GCN Layer 1:
    Units: 183
    Activation: softplus
   Duration Embedding GCN Layer 2:
    Units: 249
    Activation: gelu
    Dropout: 0.2147
 
Number of Concatenated GCN layers: 2
   Concatenated GCN Layer 1:
    Units: 190
    Skip Connections: True
    Batch Norm Momentum: 0.0161
    Batch Norm Epsilon: 2.9543e-03
    Activation: elu
    Dropout: 0.2894
   Concatenated GCN Layer 2:
    Units: 219
    Skip Connections: True
    Activation: leaky_relu
 
Pooling Method: add

Sequence Input Size: 7
 Number of Sequence Dense layers: 2
   Sequence Dense Layer 1:
    Units: 144
    Batch Norm Momentum: 0.1991
    Batch Norm Epsilon: 1.7215e-03
    Activation: gelu
   Sequence Dense Layer 2:
    Units: 134
    Activation: softplus
    Dropout: 0.3024
  
Number of Dense layers: 3
   Dense Layer 1:
    Units: 166
    Batch Norm Momentum: 0.6364
    Batch Norm Epsilon: 4.2461e-03
    Activation: tanh
   Dense Layer 2:
    Units: 252
    Batch Norm Momentum: 0.9203
    Batch Norm Epsilon: 7.9034e-03
    Activation: leaky_relu
    Dropout: 0.1207
   Dense Layer 3:
    Units: 32
    Activation: elu
    Dropout: 0.2977
 
Optimizer: RMSprop
  Learning Rate (RMSprop): 3.9195e-03
  Weight Decay (RMSprop): 3.4476e-04
  Momentum (RMSprop): 0.9454
  Alpha (RMSprop): 0.5179
  Eps (RMSprop): 6.7609e-08
  
Learning Rate Schedule: OneCycleLR
  Max_lr: 9.7978e-02
  Total_steps: 58000
  Pct_start: 0.1666

Loss function: MultiMarginLoss()
l1 lambda: 8.2976e-04
\end{lstlisting}
\end{document}